%% file: sn_news_video.tex
\documentclass[pdflatex,sn-mathphys-num]{sn-jnl}

\usepackage{graphicx}%
\usepackage{multirow}%
\usepackage{amsmath,amssymb,amsfonts}%
\usepackage{amsthm}%
\usepackage{mathrsfs}%
\usepackage[title]{appendix}%
\usepackage{xcolor}%
\usepackage{textcomp}%
\usepackage{manyfoot}%
\usepackage{booktabs}%

\usepackage{algorithm}%
\usepackage{algorithmicx}%
\usepackage{algpseudocode}%
\usepackage{listings}%
\usepackage{float}
\usepackage{makecell}

\theoremstyle{thmstyleone}%
\theoremstyle{thmstyletwo}%

\theoremstyle{thmstylethree}%

\begin{document}

\title[Article Title]{\textbf{ \large A Benchmarking Methodology  to Assess Open-Source Video Large Language Models in Automatic Captioning of News Videos}}


\author[1]{\fnm{David} \sur{Miranda Paredes}}\email{david.miranda@deepcvl.ai}

\author*[1]{\fnm{Jose M.} \sur{Saavedra}}\email{jmsaavedrar@uandes.cl}

\author[1]{\fnm{Marcelo} \sur{Pizarro}}\email{marcelo.pizarro@deepcvl.ai}

\affil*[1]{\orgdiv{Facultad de Ingeniería y Ciencias Aplicadas}, \orgname{Universidad de los Andes, Chile}, \orgaddress{\street{Av. Monseñor Álvaro del Portillo 12455}, \city{Santiago}, \postcode{100190}, \state{RM}, \country{Chile}}}




\abstract{News videos are among the most prevalent content types produced by television stations and online streaming platforms, yet generating textual descriptions to facilitate indexing and retrieval largely remains a manual process. Video Large Language Models (VidLLMs) offer significant potential to automate this task, but a comprehensive evaluation in the news domain is still lacking. This work presents a comparative study of eight state-of-the-art open-source VidLLMs for automatic news video captioning, evaluated on two complementary benchmark datasets: a Chilean TV news corpus (approximately 1,345 clips) and a BBC News corpus (9,838 clips). We employ lexical metrics (METEOR, ROUGE-L), semantic metrics (BERTScore, CLIPScore, Text Similarity, Mean Reciprocal Rank), and two novel fidelity metrics proposed in this work: the Thematic Fidelity Score (TFS) and Entity Fidelity Score (EFS). Our analysis reveals that standard metrics exhibit limited discriminative power for news video captioning due to surface-form dependence, static-frame insensitivity, and function-word inflation. TFS and EFS address these gaps by directly assessing thematic structure preservation and named-entity coverage in the generated captions. Results show that Gemma~3 achieves the highest overall performance across both datasets and most evaluation dimensions, with Qwen-VL as a consistent runner-up.}




\maketitle

\input{01_introduction}

\input{02_related}

\input{03_methodology}
\input{04_experiments}
\input{05_conclusions}

\backmatter








\section*{Declarations}

\begin{itemize}
\item \textbf{Funding}:
Partially funded by the Chilean Agency of Research and Development (ANID) through the project number 1260784.

\item \textbf{Conflict of interest/Competing interests} (check journal-specific guidelines for which heading to use): Not applicable,
\item \textbf{Ethics approval and consent to participate}: Not applicable,
\item \textbf{Consent for publication}: Not applicable.
\item \textbf{Data availability}: We will make the datasets publicly available after acceptance.   
\item \textbf{Material availability}: Not applicable.
\item \textbf{Code availability}: We will make the experimentation code publicly available after acceptance.  
\item \textbf{Author contribution}:
The first two authors contributed equally in designing the methodological protocol and defining the experimental evaluation. The third author was responsible for running some experiments. 
\end{itemize}

\noindent

\bibliography{references}

\end{document}

%% file: 01_introduction.tex
\section{Introduction}
\label{sec:introduction}
We are undergoing a massive explosion of machine-learning models capable of accurately describing text, images and audio \cite{oquab:2024:oquab,chen:2023:beats,yenduri:2024:gpt}. We witness how big companies frequently release models that improve upon the previous version. For instance, we have now many LLM models available that are being leveraged in different industries, from education \cite{Parker2025} to biomedicine \cite{Sahoo2024-pc}.

On the other hand, videos have become a very relevant type of media content. We consume videos from various platforms, including online streaming services, television programs, and television news, turning video content into a ubiquitous,  highly engaging, and information-rich medium. Furthermore, we could leverage video understanding methods in applications such as surveillance videos \cite{desilva:2024}, entertainment \cite{chen:2024:video}, and autonomous driving \cite{Lee:2022}. However, due to their multimodal nature and the abundance of information, video understanding remains a very challenging problem.

Video captioning (a.k.a. Video-to-Caption) is the task of automatically generating textual descriptions from video content. This process leverages machine learning algorithms and models to analyze video frames, extract relevant features, and then produce text that accurately and coherently describes the visual content. For the last task, Large Language Models (LLMs) play a crucial role in converting video embeddings to descriptive texts. Thus, following the success of LLMs, Video Large Language Models (VidLLM) \cite{tang:2025} are becoming prevalent for diverse video-understanding tasks. A VidLLM combines the capabilities of LLMs with video analysis to understand and interact with video content. These models can analyze videos, generate text descriptions, answer questions, and even create new video content.

The significance of the Video-to-Caption task lies in its broad applications and the value it adds across various domains. For instance, this technology significantly enhances accessibility by providing textual access to information for individuals with visual impairments, thereby facilitating access to information and culture. Automatic description generation also streamlines content indexing, making videos more searchable and discoverable. In the educational sector, Video-to-Caption can be used to automatically summarize educational videos, enabling students to review content and improve learning efficiency quickly.

In this vein, news videos are among the most prevalent content types produced by online streaming or television programs \footnote{https://tvnewscheck.com/business/article/survey-95-of-respondents-believe-accessing-local-news-on-their-local-tv-station-is-important/}. News is especially challenging because it appears in uncontrolled environments that are generally crowded and noisy. In this context, a textual description of the video content is required to facilitate indexing and posterior searching, particularly for TV Stations. Although captioning is critical for understanding news video, many video industries still rely on manual tagging and description, which require significant time for large video catalogs. Thus, it is worthwhile to leverage current technologies to automatically describe and tag videos, thereby improving efficiency and productivity. However, it still lacks a fairness evaluation of VidLLM models for the described scenario, which is critical, for instance, for TV industries to make effective decisions.

Therefore, the contributions of this work follow these three axes: 

\begin{itemize}
    \item First, to deal with the lack of testing news video datasets, we propose two benchmark datasets for this context: a Chilean TV news dataset (C13) comprising approximately 1,345 clips across diverse topics, and a BBC News dataset with 9,838 clips, both annotated with editorial descriptions and thematic descriptors. 
    \item Second, we present a comprehensive comparative study of eight open-source state-of-the-art VidLLMs for news video captioning: Gemma~3 \cite{Gemma3:2025}, Vript \cite{Vript:2024}, InternVL~3.5 \cite{InternVL3:2025}, Video-LLaMA~2 \cite{VideoLLaMa2:2024}, Video-LLaMA~3 \cite{VideoLLaMa3:2025}, LLaVA-NeXT-Video \cite{LLaVANeXTVideo:2024}, LLaVA-OneVision \cite{LLaVAOneVision:2024}, and Qwen-VL \cite{QwenVL:2025}. Our extensive experimentation encompasses both lexical metrics (METEOR, ROUGE-L) and semantic metrics (BERTScore, CLIPScore, Text Similarity, Mean Reciprocal Rank). 
    
    \item Third, we introduce two novel evaluation metrics tailored for news video captioning: the Thematic Fidelity Score (TFS), which quantifies how well generated captions preserve the thematic structure of the source video through zero-shot classification, and the Entity Fidelity Score (EFS), which measures the accuracy of named entity coverage via fuzzy matching. 
\end{itemize}
Our results reveal that Gemma~3 consistently achieves the highest overall performance across both datasets, while LLaVA-OneVision excels in token-level similarity metrics such as BERTScore and ROUGE-L.

To facilitate the reading of this document, we present the following organization. Related work is presented in Section \ref{sec:related}. A complete description of our experimental methodology is presented in Section \ref{sec:methodology}. Experiments and results are discussed in Section \ref{sec:results}. In the end, Section \ref{sec:conclusions} describes our final remarks and conclusions.

%% file: 02_related.tex
\section{Related Work}
\label{sec:related}

\subsection{Video Description}
Before delving into existing methodologies and models, it's crucial to grasp the general idea behind most Video-to-Text models. A typical pipeline for video captioning, as conceptually represented in many foundational works, begins with a raw video input (e.g., in .mp4 format). This raw video must first be transformed into a sequence of images. This is achieved by intelligently extracting representative frames from the video. Subsequently, for a computer to interpret these images, they must be converted into numerical features. These features are essentially a numerical interpretation of the image content. Therefore, the next step is to extract these features from each selected frame. This information is then fed into a text-generative model to produce corresponding captions (subtitles, comments, descriptions). Finally, these individual captions are reassembled into a coherent sequence, taking into account the original video's temporal structure.

\begin{figure}[ht!]
    \centering
    \includegraphics[width=\linewidth]{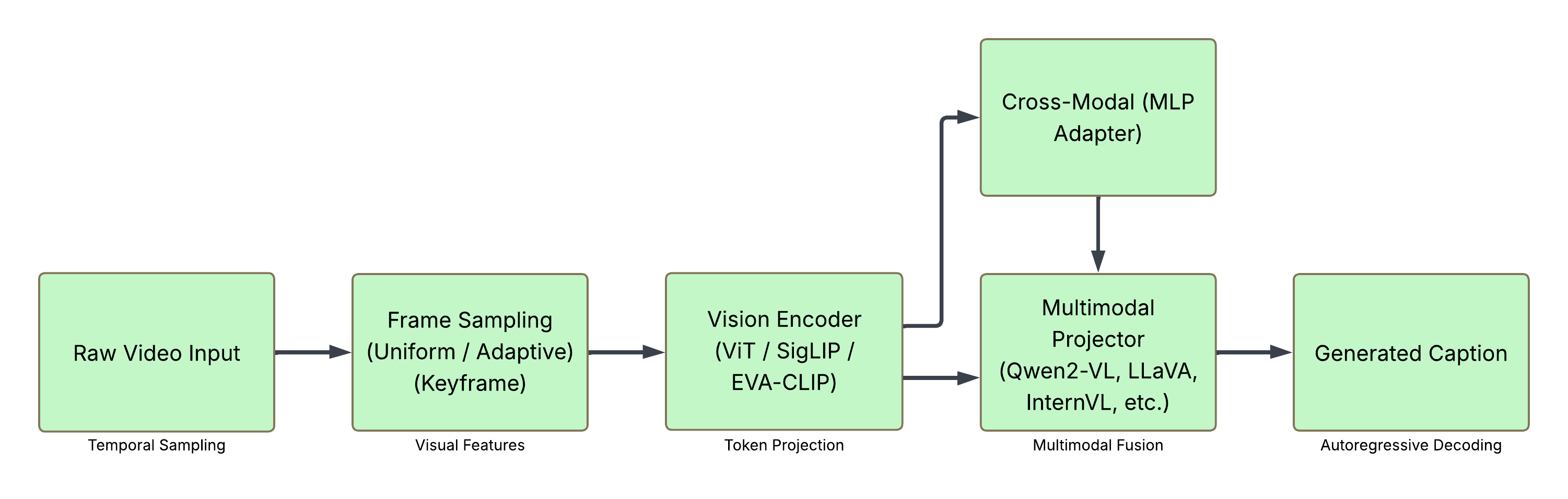}
    \caption{General pipeline of modern video captioning systems based on vision-language models (VLMs). The process begins with the raw video input, which undergoes temporal frame sampling (uniform, adaptive, or keyframe-based). The selected frames are processed by a vision encoder (typically ViT, SigLIP, EVA-CLIP, or similar architectures) to extract visual features. These features are then mapped into the language model's embedding space via a cross-modal projector (usually an MLP adapter). Finally, the multimodal LLM (such as Qwen2-VL, LLaVA-series, InternVL, among others) integrates the visual tokens with the textual prompt and autoregressively generates the natural language caption.}
    \label{fig:caption_pipeline}
\end{figure}

It's important to note that this seemingly straightforward pipeline presents several critical design choices and inflection points. These include, but are not limited to, determining the most informative key frames, the specific methods used for feature extraction, how captions are reassembled to maintain temporal coherence with the original video, and the architecture of the generative model. Consequently, the specific implementation of the Video-to-Text process can vary significantly, depending on these numerous decisions, and thus yield diverse results.

\subsection{Video Description Models}

This study examines open-source models, as open sourcing fosters innovation by enabling code inspection, modification, and improvement, resulting in faster bug fixes, higher quality, and customized solutions. Thus, the selected video description models span the current landscape of AI capabilities in multimodal understanding. These eight models employ diverse architectural designs and training paradigms to generate textual descriptions from video content. As detailed in Tables~\ref{tab:model_authors} and~\ref{tab:model_architectures}, they represent a spectrum of open-source approaches ranging from lightweight, efficient models to more parameter-rich variants, with release dates from 2024--2025.

The models fall into two primary categories. The first comprises vision-language models that integrate visual encoding with large language models, leveraging various vision encoders (SigLIP, InternViT, CLIP ViT, EVA ViT) and language backbones (Gemma, Qwen2.5, Mistral). These models differ in their approaches to temporal aggregation and frame processing, ranging from simple frame sampling to more sophisticated mechanisms such as temporal pooling and frame pruning. The second category includes models that incorporate audio alongside visual content, enabling multimodal processing of video with soundtrack or narration, which is particularly relevant for television news broadcasts.

\noindent
\begin{table}[ht!]
    \resizebox{\linewidth}{!}{    
    \centering
    \begin{tabular}{l|c|c}
        & \textbf{Author} & \textbf{Year} \\ 
        \hline
        Gemma 3.0 & Google DeepMind & 2025 \\
        Vript & Yang et al., Shanghai Jiao Tong Univ. & 2024 \\
        InternVL 3.5 & OpenGVLab, Shanghai AI Lab & 2025 \\ 
        VideoLLaMA 2 & Alibaba DAMO Academy & 2024 \\
        VideoLLaMA 3 & Alibaba DAMO Academy & 2025 \\
        LLaVA-NeXT-Video & Li et al., ByteDance / NTU & 2024 \\ 
        LLaVA-OneVision & Li et al., ByteDance / NTU & 2024 \\
        Qwen-VL & Alibaba Cloud, Qwen Team & 2025 \\ 
        \hline
    \end{tabular}
    }
    \caption{Author and release year of each video description model evaluated in this study.}
    \label{tab:model_authors}    
    
\end{table}

\noindent
\begin{table}[ht!]
    \resizebox{\linewidth}{!}{
    \centering
    \begin{tabular}{l|c|c|c|c}
    Model & \textbf{Architecture} & \textbf{\makecell{Visual\\ Encoder}} & \textbf{\makecell{LLM \\ Backbone}}  & \textbf{Size (Parameters)}\\ 
    \hline
        Gemma 3.0 & \makecell{SigLIP encoder +\\  MLP projector} & SigLIP & Gemma & Gemma-3-12b-it \\
        Vript & \makecell{Video-script alignment +\\ GPT-4V captioning} & EVA ViT-g & ST-LLM (Vicuna) & Vriptor-STLLM-7B \\
        InternVL 3.5 & \makecell{InternViT + \\MLP projector} & InternViT-6B & Qwen2.5 & InternVL3-8B \\ 
        VideoLLaMA 2 & \makecell{STC + \\BEATs Audio encoder +\\ ViT + Mistral} & CLIP ViT-L/14 & Mistral-Instruct, Qwen2 & VideoLLaMA2.1-7B \\
        VideoLLaMA 3 & \makecell{STC + \\ Differential frame pruner + \\ SigLIP} & SigLIP & Qwen2.5-7B & VideoLLaMA3-7B \\
        LLaVA-NeXT-Video & \makecell{AnyRes +\\ temporal pooling +\\ MLP projector} & CLIP ViT-L/14 & Qwen2 & LLaVA-NeXT-Video-7B \\ 
        LLaVA-OneVision & \makecell{AnyRes + \\ multi-image curriculum} & SigLIP & Qwen2 & LLaVA-OneVision-7B \\ 
        Qwen-VL & ViT + \makecell{Multimodal RoPE + \\ token merging} & ViT-600M & Qwen2.5 & Qwen2.5-VL-7B \\ 
    \end{tabular}
    }
    \caption{Architectural details and parameter counts of each video description model evaluated in this study.}
    \label{tab:model_architectures}
    
\end{table}

%% file: 03_methodology.tex
\section{Methodology}
\label{sec:methodology}
        
\subsection{Loading Models}

Loading the selected models, described in Table \ref{tab:model_architectures}, requires a careful, case-specific process, given the diversity of architectures, dependencies, and distribution methods available. Despite these differences, a common methodology was established to ensure correct initialization, inference execution, and result storage, which is described below.

\begin{itemize}
\item \textbf{Gemma 3}: As an open model released to the community by Google DeepMind, Gemma 3.0's preparation was streamlined. Its weights and necessary tokenizer files were readily available on prominent model hubs, specifically HuggingFace. 
To use the model's capabilities to the fullest, a fixed number of extracted frames had to be calculated in advance (manually) to avoid filling the input context window.

\item \textbf{Vript}: This model requires a more complex preparation process. Multiple components were downloaded and integrated: a pre-trained image encoder (EVA ViT-g), a Q-Former with a linear projection associated with InstructBLIP, and, finally, the weights of Vicuna, which served as the base for the Vriptor-STLLM model's inference. All dependencies were obtained from the official Vript project repositories, following the developers' instructions to ensure correct functionality.
    
\item \textbf{InternVL 3.5}: This model was obtained from the OpenGVLab repositories hosted on HuggingFace, using the 8B-parameter variant (\texttt{InternVL3-8B}). Since InternVL 3.5 uses InternViT-6B as its vision encoder with native dynamic-resolution support, the input frames were fed without manual resizing, allowing the model's pixel-shuffle downsampling to handle arbitrary aspect ratios. Inference was executed across two GPUs to accommodate the model's memory footprint, and the resulting captions from both partitions were merged into a single dataset keyed by clip identifier.
    
\item \textbf{VideoLLaMA2}: Preparation focused on downloading the ``safetensors'' file containing the model weights, along with complementary ``vocab.json'' and ``tokenizer\_config.json'' files, which are essential for textual input processing. The download was performed directly from HuggingFace, a platform that offered both a user-friendly interface and greater stability for integration with the developed inference scripts, enabling quick model setup and facilitating its reuse in multiple experiments.

\item \textbf{VideoLLaMA3}: Similar to its predecessor, integrating this involved downloading its model weights and associated configuration files from its official open-source repositories or platforms such as HuggingFace. This approach ensured access to the latest model version and facilitated its integration into existing inference pipelines, leveraging its vision-centric design for image and video understanding.

 \item \textbf{LLaVA-NeXT-Video}: The pre-trained weights for LLaVA-NeXT-Video were obtained from the official ByteDance/NTU release on HuggingFace. This model extends LLaVA-NeXT to the video domain by incorporating a temporal pooling mechanism over its AnyRes visual features, using CLIP ViT-L/14 as visual encoder and Qwen2 as language backbone. The model was loaded via the \texttt{transformers} library, and inference was performed by uniformly sampling frames from each clip and passing them through the two-layer MLP projector into the language model.

\item \textbf{LLaVA-OneVision}: Weights were downloaded from the official HuggingFace repository. LLaVA-OneVision is a unified multi-modal model supporting images, multi-image, and video inputs through a single architecture built on a SigLIP vision encoder and Qwen2 language backbone. For video inference, frames were uniformly sampled from each clip and processed through its image-to-video task-transfer curriculum. The model was loaded and executed using the \texttt{transformers} library with standard configuration.

\item \textbf{Qwen-VL}: The Qwen2.5-VL model was obtained from the Alibaba Cloud Qwen team's HuggingFace repository. This model features a Vision Transformer encoder with approximately 600M parameters and employs Multimodal Rotary Position Embedding (M-RoPE) to jointly encode spatial and temporal information, enabling native variable-length video processing without requiring a fixed frame count. Token merging compression reduces the visual token sequence before feeding it into the language model. Inference was performed via the \texttt{transformers} library, leveraging the model's ability to process video clips of varying durations. directly.
    
\end{itemize}

\subsection{Evaluation Metrics}
We adopted a multifaceted evaluation approach to thoroughly assess the quality of automatically generated video descriptions, leveraging several widely recognized natural language processing metrics. Each metric offers a distinct perspective on the alignment and quality of the generated text compared to human-authored reference descriptions, providing a comprehensive understanding of model performance.

\begin{itemize}
\item \textbf{ROUGE (Recall-Oriented Understudy for Gisting Evaluation)} \cite{ROUGE:2004}: Complementing precision-oriented n-gram metrics, ROUGE focuses on recall, quantifying how much of the information present in the reference descriptions is captured by the generated text. This metric is crucial for evaluating the completeness of the descriptions, penalizing omissions of important content. ROUGE-L, which measures the length of the longest common subsequence, is useful for assessing overall content consistency and fluency.

\item \textbf{METEOR (Metric for Evaluation of Translation with Explicit Ordering)} \cite{METEOR:2005}: Designed to overcome some limitations of n-gram-based metrics, METEOR provides a more nuanced evaluation by considering synonymy and paraphrasing through linguistic resources like WordNet. It assesses word correspondence and fluency, offering a more accurate judgment of description quality, especially when the generated text uses valid semantic variations of the reference.

\item \textbf{BERTScore (Precision, Recall, F1)} \cite{BERTScore:2019}: This metric delves deeper into semantic similarity by utilizing contextual embeddings from BERT models. BERTScore measures how closely the meaning of the generated text aligns with the reference, rather than just word-for-word matches. It provides Precision, Recall, and F1 scores, offering detailed insights into the relevance (Precision) and comprehensiveness (Recall) of the generated descriptions, which are vital for assessing the true semantic fidelity of the model's output. 
    
\item \textbf{CLIPScore} \cite{CLIPScore:2021}: This metric aims to evaluate the quality of the generated description with respect to the video content, leveraging the CLIP encoder's capacity to align text and image representations. Thus, we produce visual embeddings from different parts of the video that are then compared with the text embedding computed from the inferred description. For comparison, we use cosine similarity. 

\item \textbf{Text Similarity}: Beyond specific n-gram or token matches, Text Similarity is employed to assess the overall semantic similarity between the vector representations of the generated and reference captions. Both texts are encoded into 384-dimensional sentence embeddings using the \texttt{all-MiniLM-L12-v2} model from Sentence-Transformers \cite{SentenceTransformers:2019}, a lightweight distilled transformer fine-tuned on over one billion sentence pairs for semantic textual similarity.
    
\item \textbf{Reciprocal Ranking (RR)}: Added retrospectively to validate hypotheses about model performance, Reciprocal Ranking is a metric primarily used for evaluating ranked lists of results. For each clip, the reference description is compared against the pool of all model-generated captions via text similarity computed in the same \texttt{all-MiniLM-L12-v2} embedding space described above. The models are then ranked by similarity, and the reciprocal of the rank of the correct (highest-scoring) model gives the RR for that clip. The Mean Reciprocal Rank (MRR) across all clips provides an aggregate performance measure that indicates the system's effectiveness at prioritizing relevant content and can be particularly insightful for confirming a model's overall superiority across evaluation instances.
\end{itemize}

\subsubsection{Thematic Fidelity Score (TFS)}
\label{sec:tfs_metric}

While the metrics described above capture lexical, semantic, and embedding-level similarity between generated and reference captions, they do not directly assess whether the caption preserves the \emph{thematic content} of the original video. To address this gap, we propose the Thematic Fidelity Score (TFS), a novel metric that evaluates how faithfully a generated caption reflects the high-level topics present in the ground-truth description.

\paragraph{Zero-shot thematic classification.}
The first stage assigns thematic labels to both ground-truth descriptions and model generated captions. We employ a zero-shot multi-label classification pipeline using \texttt{DeBERTa-v3-base-zeroshot-v2.0} \cite{DeBERTa:2021}, a natural language inference model that can classify text against arbitrary label sets without task specific finetuning. Each text is evaluated against a set of 15 candidate thematic labels (listed in Table~\ref{tab:theme_labels}), and a label is assigned when the entailment score exceeds a threshold of $\tau = 0.5$. This produces a binary label vector $\mathbf{y} \in \{0,1\}^{15}$ for each clip.

\begin{table}[h]
\centering
\caption{The 15 candidate thematic labels used for zero-shot classification.}
\label{tab:theme_labels}
\small
\begin{tabular}{ll}
\# & Thematic Label \\
1 & Politics and Elections \\
2 & International Affairs and Conflicts \\
3 & Economy, Business and Finance \\
4 & Society and Social Issues \\
5 & Health and Medicine \\
6 & Crime and Justice \\
7 & Environment and Climate \\
8 & Science and Technology \\
9 & Arts, Culture and Entertainment \\
10 & Sports and Athletics \\
11 & Education and Academia \\
12 & Natural Disasters and Weather \\
13 & Accidents and Emergencies \\
14 & Animals and Wildlife \\
15 & History and Heritage \\
\end{tabular}
\end{table}

\paragraph{Thematic Fidelity Score.}
Let $\mathbf{y}^{gt}$ and $\mathbf{y}^{pred}$ denote the ground-truth and predicted binary label vectors, respectively. For each clip $i$ and model $m$, the Thematic Fidelity Score is defined as the micro-averaged F1-score:

\begin{equation}
\label{eq:tfs}
\text{TFS}_{i,m} = \text{T\text{}F1}_{i,m} = \frac{2\,\text{TP}}{2\,\text{TP} + \text{FP} + \text{FN}}
\end{equation}

\noindent where TP (true positive), FP (false positive), and FN (false negative) are computed from the element-wise comparison of $\mathbf{y}^{gt}$ and $\mathbf{y}^{pred}$ across all 15 themes for a single clip. This score captures thematic classification accuracy per clip, rewarding models that correctly identify the themes present in the ground-truth while penalizing both missed themes (FN) and hallucinated themes (FP) through the F1 formulation.

\subsubsection{Entity Fidelity Score (EFS)}
\label{sec:efs_metric}

Complementing the thematic perspective of TFS, the Entity Fidelity Score (EFS) evaluates how well a generated caption recovers specific named entities persons, locations, organizations, and geopolitical entities mentioned in the ground-truth description. Named entities are critical anchors in news descriptions, and their faithful reproduction is a strong indicator of factual grounding.

\paragraph{Named entity extraction}
Both ground-truth descriptions and model-generated captions are processed through spaCy's \texttt{en\_core\_web\_sm} NER pipeline, configured to extract entities of types \texttt{PERSON}, \texttt{GPE}, \texttt{ORG}, \texttt{LOC}, \texttt{NORP}, \texttt{FAC}, and \texttt{EVENT}. All extracted entity strings are normalized to lowercase with collapsed whitespace.

\paragraph{Fuzzy matching}
Because models may produce slight variations of entity names (e.g., ``Boric'' vs.\ ``Gabriel Boric''), we employ fuzzy string matching via the RapidFuzz library. Two entity strings are considered a match when their token-ratio similarity exceeds a threshold of $\theta = 85$. This allows credit for partial yet recognisable entity mentions.

\paragraph{EFS formula}
For each clip $i$ and model $m$, let $E^{gt}_i$ and $E^{m}_i$ denote the sets of entities extracted from the ground-truth and model caption, respectively. Entity matching is performed via the fuzzy comparison described above, yielding a set of matched ground-truth entities $M^{gt}_i \subseteq E^{gt}_i$ and a set of matched model entities $M^{m}_i \subseteq E^{m}_i$. The Entity Fidelity Score is then defined as the harmonic mean of entity precision and recall:

\begin{equation}
\label{eq:efs}
\text{EFS}_{i,m} = \frac{2 \times \text{Precision}_{i,m} \times \text{Recall}_{i,m}}{\text{Precision}_{i,m} + \text{Recall}_{i,m}}
\end{equation}

\noindent where:
\begin{itemize}
    \item $\text{Precision} = \frac{|M^{m}_i|}{\max(|E^{m}_i|, 1)}$: the fraction of model-predicted entities that match a ground-truth entity, penalising hallucinated entities.
    \item $\text{Recall} = \frac{|M^{gt}_i|}{\max(|E^{gt}_i|, 1)}$: the fraction of ground-truth entities recovered by the model caption.
\end{itemize}

\noindent When a clip contains no ground-truth entities ($|E^{gt}_i| = 0$), the clip is excluded from the EFS evaluation, as entity fidelity cannot be meaningfully assessed.

\subsection{Datasets}
We aim to evaluate the selected models in terms of their capabilities to describe television news in uncontrolled environments. To this end, we propose two news video datasets, collected from an important local television station (C13) and from an international broadcasting company (BBC). Following, we describe each of the proposed datasets.

\subsubsection{C13 (ChTv)}
This dataset was obtained through an internal collaboration with a local television station, as part of a cutting-edge knowledge exchange program. It comprises approximately 100 videos, ranging in duration from 1 to 3 hours, corresponding to the channel's morning programs and newscasts. The significant variability in both length and content of these videos allowed for evaluating the models' performance under more realistic and challenging conditions. In addition to the raw video files, the local station provided JSON files containing manual tags and descriptions for most of the programs. These annotations, generated on their existing editing platform by their team of librarians, served as a crucial ``reference tagging’’ for comparative evaluation of the models' performance. Subsequently, approximately 25 GB of audiovisual material from the TV station was received. As previously mentioned, a JSON file was also attached, containing relevant information for each long video, including duration, start timestamp, description, title, and other attributes. 

Then we split long videos into individual clips using FFmpeg and other video libraries. These clips, corresponding to each annotation, were extracted based on their ``start'' and ``duration'' values. The extraction was limited to a duration of at least 10 seconds and no more than 300 seconds (5 minutes). This constraint prevented evaluations that could overload GPU memory and avoided assessments on very short clips that might yield insufficient context for meaningful understanding. Through this process, we build a dataset of 1345 video clips, each paired with a human-generated descriptive annotation and referential information, enabling subsequent experimentation and evaluation.

\begin{figure}[h!]
  \centering
  \includegraphics[width=0.6\linewidth]{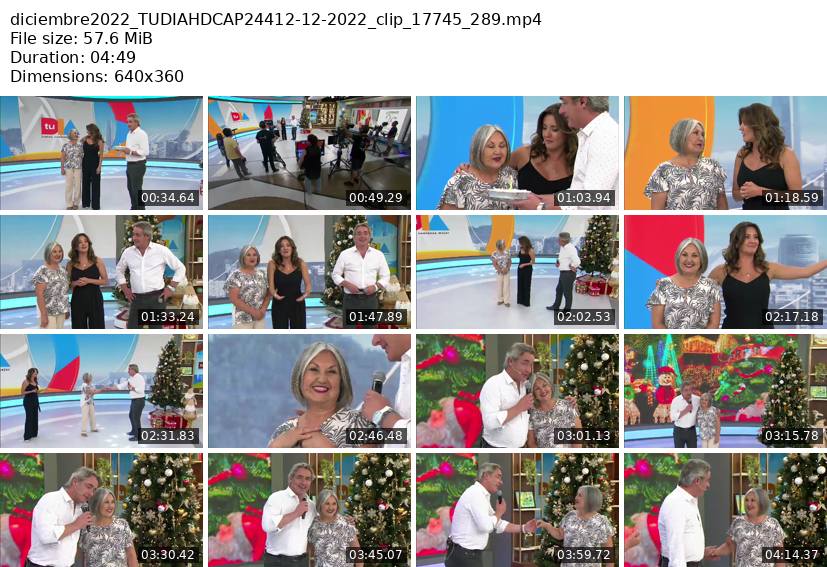}
    \includegraphics[width=0.6\linewidth]{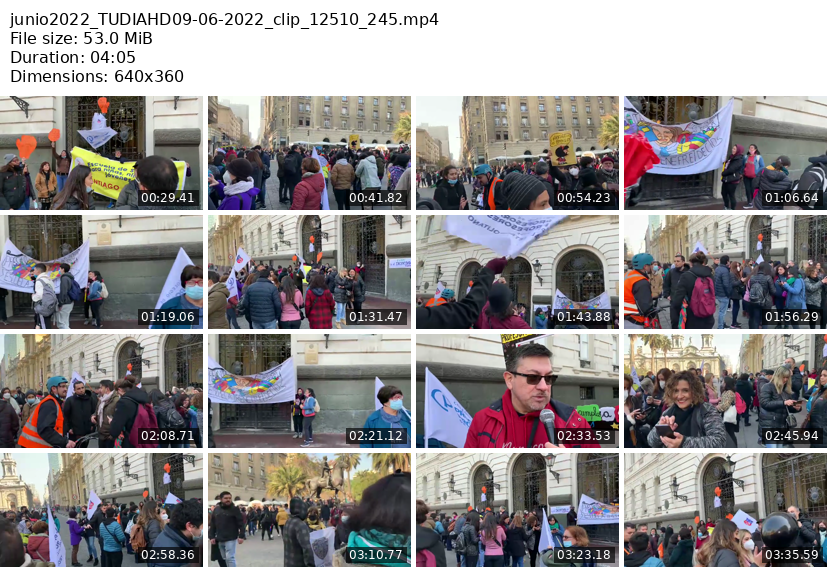}
  \caption{A sample of video content from the ChTv dataset. ChTv was downloaded from \url{https://www.13.cl/}.}
  \label{fig:chtv_example}
\end{figure}

The ChTv dataset provides a diverse range of content, such as news segments and morning shows, as exemplified in Figure~\ref{fig:chtv_example}. The variety in content ensures that the models are tested under conditions that reflect real-world use cases for video understanding systems.

\begin{figure}[h!]
  \centering
  \includegraphics[width=\linewidth]{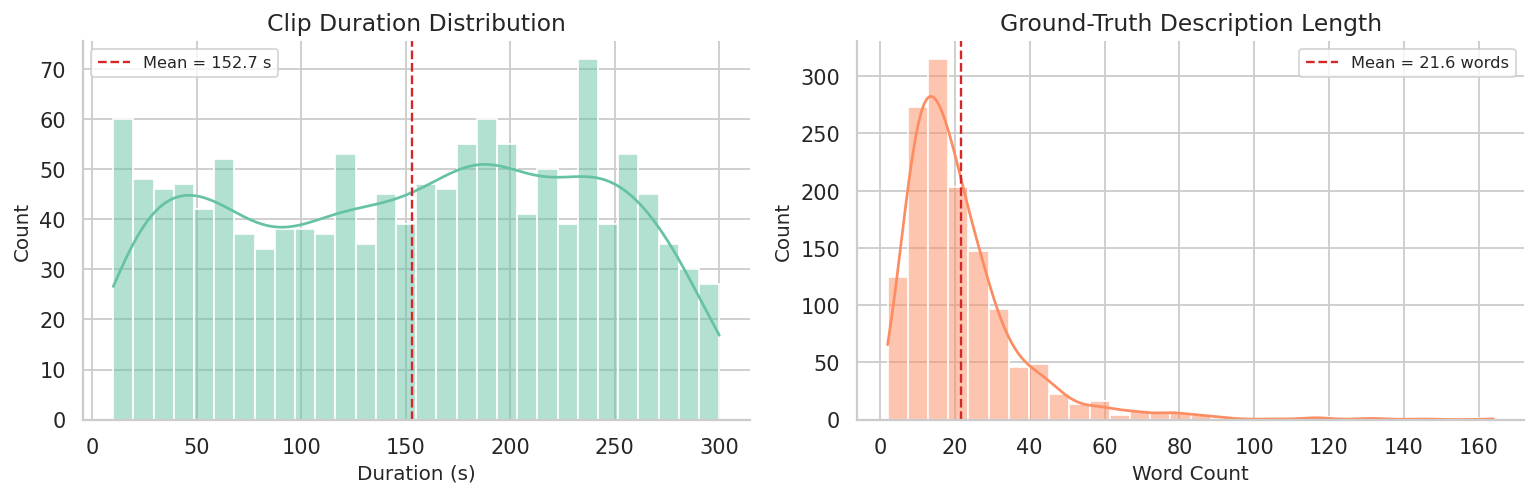}
  \caption{Distribution of Video Clip durations and their description lengths (number of words) within the ChTv Dataset.}
  \label{fig:chtv_duration_length}
\end{figure}

As shown in Figure~\ref{fig:chtv_duration_length}, the distribution of the video clip durations in the ChTv dataset varies from short clips around 10 seconds to longer clips that can extend to 5 minutes. This distribution enables us to evaluate how well the models can handle clips of varying durations, from brief segments to those requiring more in-depth analysis.

The description lengths associated with each video clip are shown in Figure~\ref{fig:chtv_duration_length}. As indicated, most descriptions are fewer than 40 words, while some clips feature more detailed descriptions. This variation in description length adds another layer of complexity when evaluating the models' ability to process and understand textual annotations.

\begin{figure}[h!]
  \centering
  \includegraphics[width=0.8\linewidth]{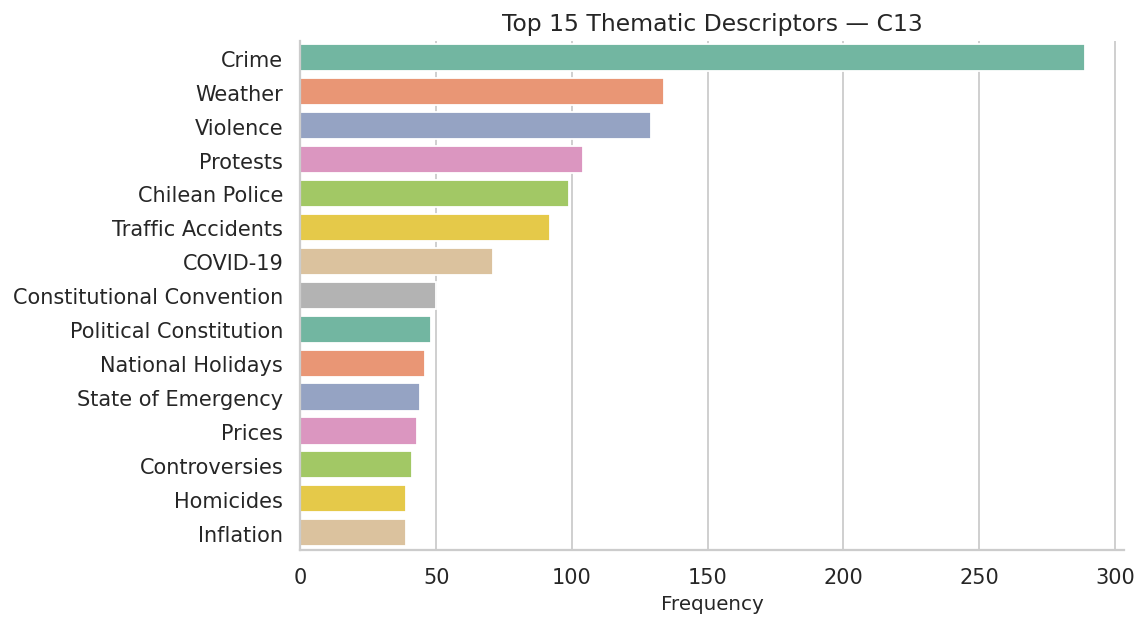}
  \caption{Distribution of 'Thematic Descriptors' in the ChTv Dataset.}
  \label{fig:chtv_thematics_descriptors}
\end{figure}

Figure~\ref{fig:chtv_thematics_descriptors} displays the distribution of thematic descriptors within the dataset, translated from Spanish to English using a curated dictionary of over 200 tag mappings. The most frequent thematic descriptors include ``Crime'', ``Weather'' and ``Violence''. This variety of thematic descriptors reflects the diverse nature of the content covered in the dataset, which spans both news related topics and broader societal themes. These descriptors also help to define the ground-truth thematic labels for the Thematic Fidelity Score (TFS) described in Section~\ref{sec:tfs_metric}.

\subsubsection{BBC News} 
This dataset was assembled from publicly available broadcast news content collected through a large-scale automated curation process. The source material was identified using sitemap indexes from the BBC website, obtained via the site’s robots.txt file, which provides structured references to archived video pages. These sitemaps enabled the systematic discovery of thousands of news entries containing embedded video segments and associated editorial metadata. From this source, a large corpus of audiovisual material was gathered and filtered to retain clips with manageable durations for computational analysis.

\begin{figure}[ht]
  \centering
  \includegraphics[width=0.6\linewidth]{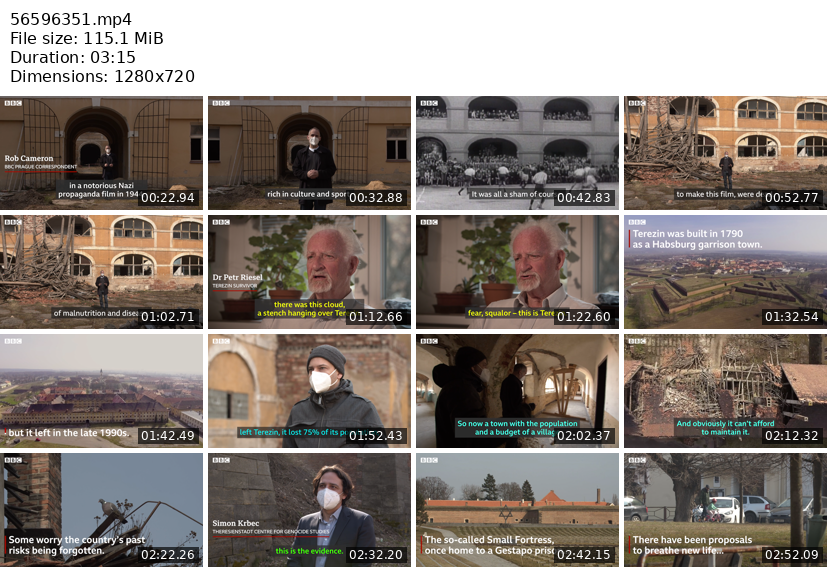}
  \includegraphics[width=0.6\linewidth]{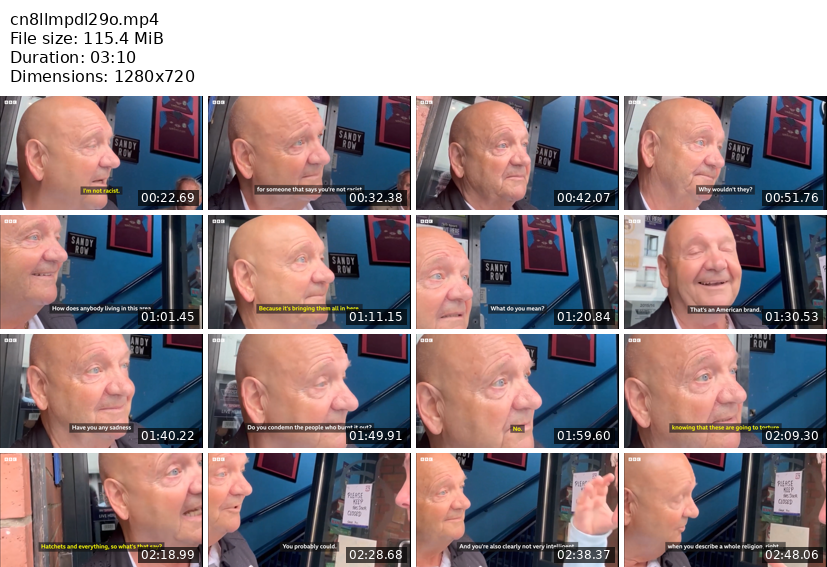}
  \caption{A sample of video content from the BBC dataset. BBC was downloaded from \url{https://www.bbc.com/}.}
  \label{fig:bbc_example}
\end{figure}

The final dataset comprises 9,838 videos, each with a duration not exceeding 300 seconds, amounting to approximately 370 GB of data. This upper bound on clip length ensured consistency across samples and prevented excessive memory requirements during model training and evaluation, while still preserving sufficient temporal context for meaningful understanding. The content spans a wide range of news topics, visual scenarios, and production styles, providing significant variability and enabling the evaluation of models under realistic and heterogeneous conditions. Figure \ref{fig:bbc_example} depicts a sample of the proposed BBC News dataset.

\begin{figure}[h!]
  \centering
  \includegraphics[width=\linewidth]{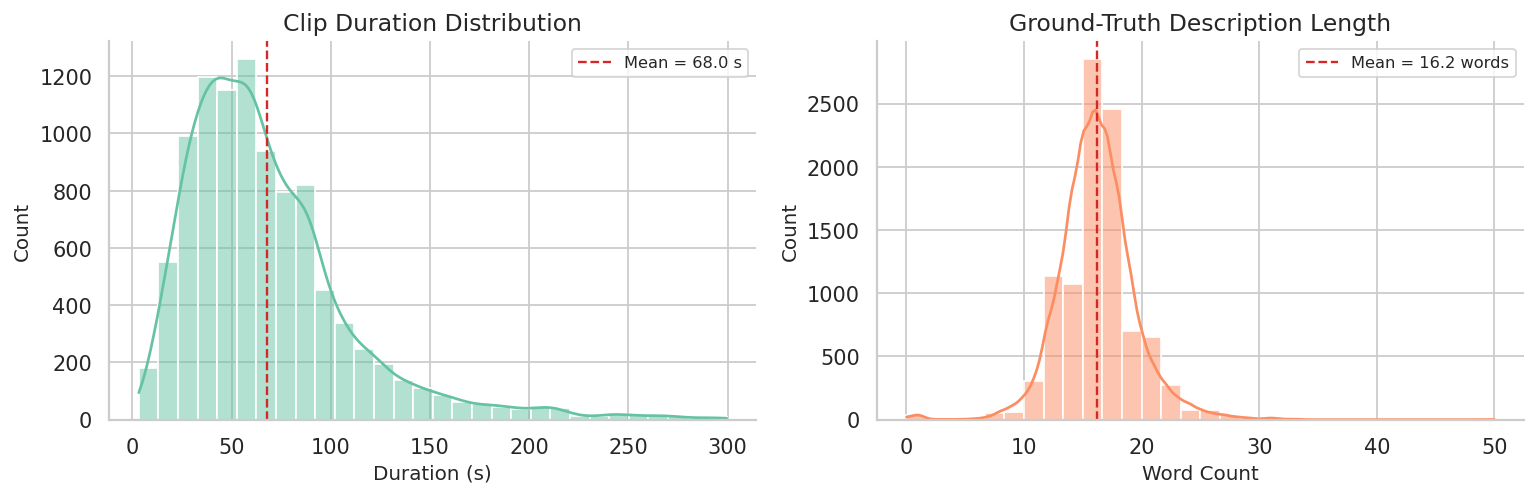}
  \caption{Distribution of Video Clip durations and their description lengths (number of words) within the BBC News Dataset.}
  \label{fig:bbc_duration_length}
\end{figure}

\begin{figure}[h!]
  \centering
  \includegraphics[width=0.8\linewidth]{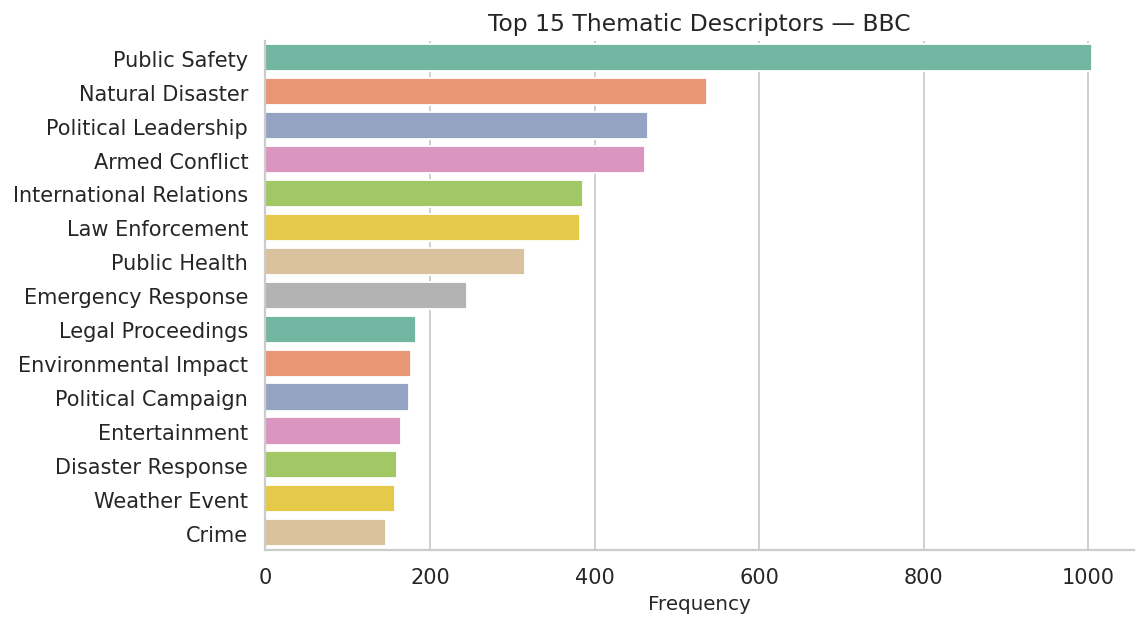}
  \caption{Distribution of 'Thematic Descriptors' in the BBC News Dataset}
  \label{fig:bbc_thematics_descriptors}
\end{figure}

As shown in Figure~\ref{fig:bbc_duration_length}, the distribution of clip durations in the BBC dataset is concentrated in shorter segments, with most clips lasting under 180 seconds. The description lengths exhibit a left distribution, with the majority of editorial descriptions containing fewer than 50 words. Figure~\ref{fig:bbc_thematics_descriptors} presents the thematic descriptor distribution, revealing a broad coverage of news topics that enables cross-dataset comparison with the ChTv corpus.

Alongside the video files, structured metadata were compiled for each clip, including duration, title, and descriptive text originally associated with the corresponding news entry. These descriptions function as weak textual annotations that allow alignment between visual and semantic content, facilitating tasks such as retrieval, multimodal representation learning, and temporal understanding. Because the annotations originate from editorial descriptions rather than manual labeling for research purposes, they reflect real-world descriptive practices and introduce a level of noise and variability that more closely resembles operational environments.

Through this filtering and aggregation process, a large-scale dataset of short news video clips paired with descriptive metadata was created. Its controlled duration range, consistent formatting, and substantial scale make it suitable for systematic experimentation and benchmarking, while the diversity of visual content and editorial descriptions supports robust evaluation of video understanding models in realistic news-media contexts.

%% file: 04_experiments.tex
\section{Experiments and Results}
\label{sec:results}
In this section, we present a comprehensive evaluation of the eight selected leading multimodal models for video description generation: Gemma~3, Vript, InternVL~3.5, Video-LLaMA~2, Video-LLaMA~3, LLaVA-NeXT-Video, LLaVA-OneVision, and Qwen-VL. Each model was evaluated on both the ChTV (Chilean TV news) and BBC News datasets using a suite of n-gram, semantic, ranking, and fidelity-based metrics.

\subsection{Evaluation Metrics and Model Performance}

\subsubsection{Overall Performance Summary}
To measure the quality of the generated descriptions, we employed a comprehensive multifaceted evaluation approach using traditional n-gram-based scores (METEOR and ROUGE-L), semantic-based evaluations (CLIPScore, BERTScore, and Text Similarity), as well as two novel fidelity metrics: the Thematic Fidelity Score (TFS) and Entity Fidelity Score (EFS). Tables~\ref{tab:c13_metrics} and~\ref{tab:bbc_metrics} present a complete summary of all model performance across both datasets and all evaluation metrics.

\begin{table}[h!]
    \resizebox{\linewidth}{!}{
    \centering
    \begin{tabular}{lccccc}
        \hline
        Model & Text Sim. & CLIPScore & METEOR & ROUGE-L & BERTScore F1 \\
        \hline
        Gemma 3           & \textbf{0.3112} & \textbf{0.3385} & \textbf{0.1281} & 0.0721 & 0.7809 \\
        Vript             & 0.2227 & 0.3175 & 0.1063 & 0.0604 & 0.7726 \\
        InternVL 3.5      & 0.2291 & 0.3178 & 0.1040 & 0.0583 & 0.7699 \\
        Video-LLaMA 2     & 0.2007 & 0.3134 & 0.1135 & 0.0768 & 0.7819 \\
        Video-LLaMA 3     & 0.1794 & 0.3065 & 0.0984 & 0.0766 & 0.7867 \\
        LLaVA-NeXT-Video  & 0.1999 & 0.3121 & 0.1050 & 0.0552 & 0.7703 \\
        LLaVA-OneVision   & 0.1996 & 0.3098 & 0.0908 & \textbf{0.0824} & \textbf{0.7905} \\
        Qwen-VL           & 0.2672 & 0.3283 & 0.1097 & 0.0575 & 0.7716 \\
        \hline
    \end{tabular}
        }
    \caption{Summary of model performance metrics on the \textbf{ChTV dataset}. Bold values indicate the best score per metric.}
    \label{tab:c13_metrics}
\end{table}

\begin{table}[h!]
    \resizebox{\linewidth}{!}{
    \centering
    \begin{tabular}{lccccc}
        \hline
        Model & Text Sim. & CLIPScore & METEOR & ROUGE-L & BERTScore F1 \\
        \hline
        Gemma 3           & \textbf{0.4387} & \textbf{0.3642} & \textbf{0.2017} & 0.1014 & 0.8330 \\
        Vript             & 0.2371 & 0.3228 & 0.1194 & 0.0624 & 0.8046 \\
        InternVL 3.5      & 0.2787 & 0.3358 & 0.1310 & 0.0669 & 0.8067 \\
        Video-LLaMA 2     & 0.2901 & 0.3237 & 0.1507 & 0.0928 & 0.8219 \\
        Video-LLaMA 3     & 0.3172 & 0.3426 & 0.1591 & 0.0874 & 0.8183 \\
        LLaVA-NeXT-Video  & 0.2294 & 0.3207 & 0.1202 & 0.0622 & 0.8048 \\
        LLaVA-OneVision   & 0.2747 & 0.3336 & 0.1363 & \textbf{0.1202} & \textbf{0.8360} \\
        Qwen-VL           & 0.3607 & 0.3543 & 0.1467 & 0.0709 & 0.8142 \\
        \hline
    \end{tabular}
    }
    \caption{Summary of model performance metrics on the \textbf{BBC dataset}. Bold values indicate the best score per metric.}
    \label{tab:bbc_metrics}
\end{table}

\paragraph{Traditional Metrics}
METEOR and ROUGE-L are established metrics for evaluating machine-generated texts, originally designed for translation and summarization, respectively. However, both assume a high degree of surface-form overlap (exact string matching) between the generated and reference texts, an assumption that is systematically wrong in news video captioning. METEOR relies on exact, stemmed, and synonym token matches against a single reference, yet models produce abstractive descriptions that rephrase editorial content rather than reproduce it verbatim. As a result, METEOR scores remain low and compressed (0.098--0.128 on ChTV; 0.119--0.202 on BBC), with too little spread to meaningfully differentiate model quality. ROUGE-L measures the longest common subsequence between two texts; when the reference is a concise editorial tag (ChTV) or a short indexing summary (BBC) and the generated caption is a detailed visual narrative, the LCS will be short regardless of caption quality. All models score below 0.09 on ChTV and below 0.16 on BBC (with the exception of Gemma~3's 0.202), leaving ROUGE-L similarly undiscriminating. These limitations confine both metrics to detecting only gross failures rather than capturing fine-grained differences in caption quality.

\paragraph{Semantic Metrics}
The semantic metrics CLIPScore, BERTScore, and Text Similarity leverage latent spaces inferred by state-of-the-art encoders and, in principle, move beyond surface form matching. In practice, however, each exhibits specific limitations in the context of news video captioning.

CLIPScore evaluates visual--textual alignment using CLIP embeddings. While well suited for photo-captioning tasks, in news television the video frames are frequently dominated by static shots of reporters addressing the camera or studio backdrops. Consequently, the frame embeddings carry limited discriminative information about the thematic or factual content being discussed, and CLIPScore reflects the general visual context rather than caption accuracy. The narrow range across models (0.306--0.339 on ChTV; 0.321--0.364 on BBC) confirms that this metric offers limited power to differentiate model quality in this domain.

BERTScore F1 performs token-level alignment using contextual BERT embeddings. Although more semantically aware than n-gram overlap, its scores are inflated by the high proportion of common function words (articles, prepositions, conjunctions, adverbs) that naturally overlap between any two coherent English texts discussing similar general topics. All models score above 0.77 on ChTV and above 0.80 on BBC, producing a ceiling effect with a spread of only 0.021 on ChTV and 0.031 on BBC, which limits meaningful differentiation.

\textbf{Text Similarity} (sentence-level cosine similarity via \texttt{all-MiniLM-L12-v2}) provides the widest spread among the five metrics (0.132 on ChTV, 0.209 on BBC) and is validated by the shuffled-pairs test described below. Nonetheless, it still measures general semantic proximity at the sentence level without assessing whether specific thematic content or named entities are correctly captured in the generated caption.

These limitations, insensitivity to thematic coverage (CLIPScore, BERTScore), dependence on surface-form overlap (METEOR, ROUGE-L), motivate the two novel fidelity metrics proposed in this work. The Thematic Fidelity Score (TFS) and Entity Fidelity Score (EFS), presented in Section ~\ref{sec:tfs_results}, are specifically designed to evaluate whether generated captions preserve the thematic structure and named entities of the source content, addressing evaluation dimensions that existing metrics leave uncovered.

\subsubsection{Validation of Semantic Evaluation}

\begin{figure}[h!]
  \centering
  \includegraphics[width=0.95\linewidth]{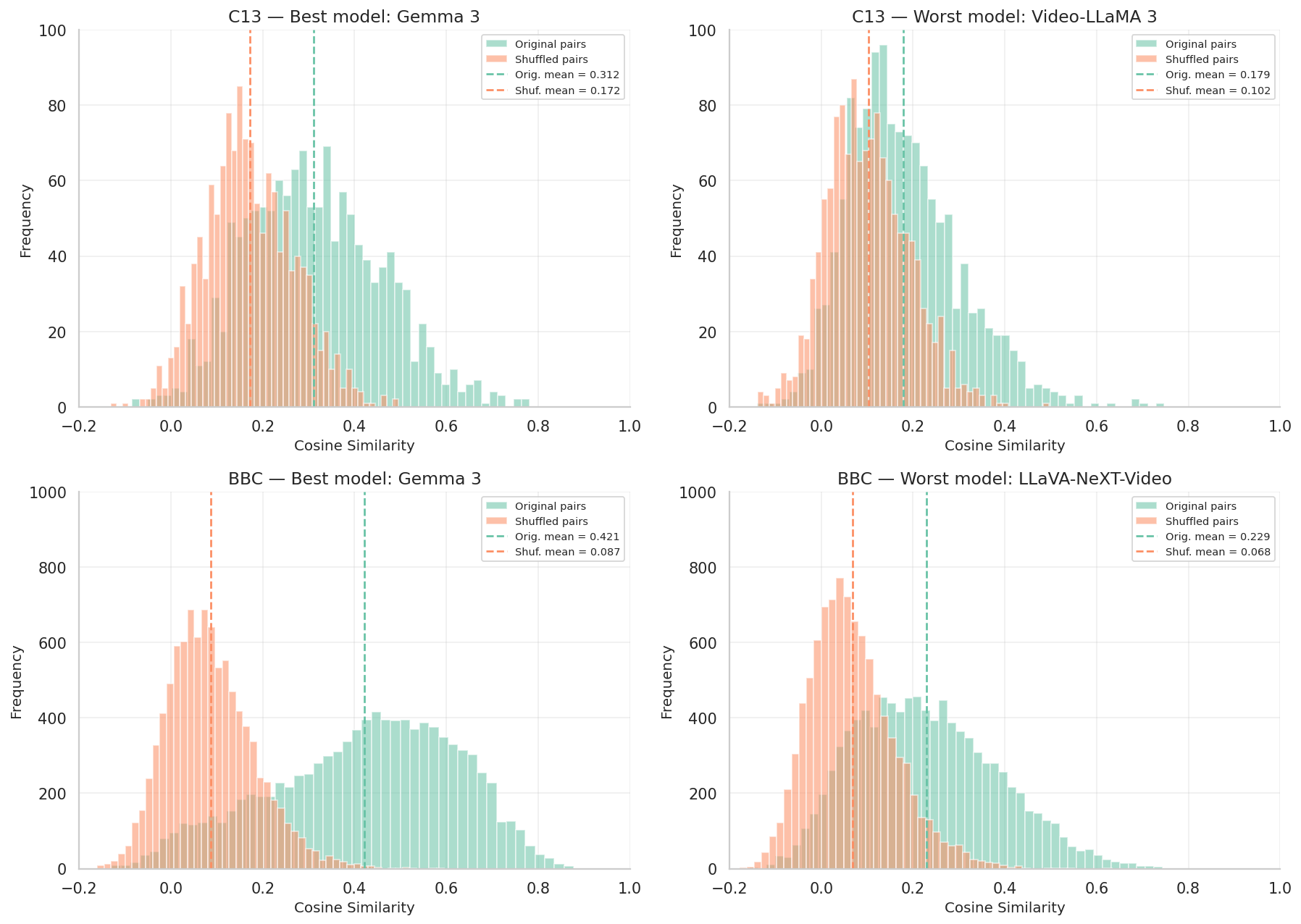}
  \caption{Shuffled-pairs validation of Text Similarity for the best-performing model (Gemma~3) and the lowest-performing model, illustrating the discriminative power of the metric.}
  \label{fig:cosine_sim_validation}
\end{figure}

To validate the effectiveness of Text Similarity as a metric in this context, we conducted a ``shuffled pairs'' test (see Figure~\ref{fig:cosine_sim_validation}). For each model, we compared the distribution of similarity scores from the original (correct) video description pairs against a distribution from shuffled pairs, where generated descriptions were randomly matched with human references from different video clips. For Gemma~3 (the best-performing model), there is a clear separation between the original and shuffled pair distributions on both datasets, confirming that Gemma~3 generates descriptions that are specifically relevant to their corresponding videos. Conversely, for lowest performing models, the distributions show considerably less separation, indicating a tendency to produce more generic descriptions. This demonstrates that low semantic similarity may reflect a fundamental mismatch between the model's descriptive style and the nature of the ground-truth data, rather than universally poor descriptive capabilities.

\subsubsection{Ranking Performance and Description Length}

\paragraph{Mean Reciprocal Rank (MRR)}

To further assess model quality, we computed the Mean Reciprocal Rank (MRR) based on Text Similarity. For each video-clip, the descriptions generated by all eight models are ranked by their similarity to the reference description. Reciprocal Rank (RR) measures the position of the most similar description in that ranked list, and MRR is the average across all clips. The results are presented in Table~\ref{tab:mrr_rankings}.

\begin{table}[h!]
    \centering
    \begin{tabular}{l|cc}
        \hline
        \textbf{Model} & \textbf{ChTV Dataset} & \textbf{BBC Dataset} \\
        \hline
        Gemma 3          & 0.619 & 0.681 \\
        Vript            & 0.319 & 0.218 \\
        InternVL 3.5     & 0.326 & 0.263 \\
        Video-LLaMA 2    & 0.269 & 0.305 \\
        Video-LLaMA 3    & 0.235 & 0.328 \\
        LLaVA-NeXT-Video & 0.252 & 0.211 \\
        LLaVA-OneVision  & 0.247 & 0.261 \\
        Qwen-VL          & 0.450 & 0.451 \\
        \hline
    \end{tabular}
    \caption{Mean Reciprocal Rank (MRR) across models and datasets. Gemma~3 achieves the highest MRR on both datasets, confirming that it consistently ranks first or near-first among all models for any given video.}
    \label{tab:mrr_rankings}
\end{table}

\paragraph{Description Length}

\begin{figure}[h!]
  \centering
  \includegraphics[width=0.8\linewidth]{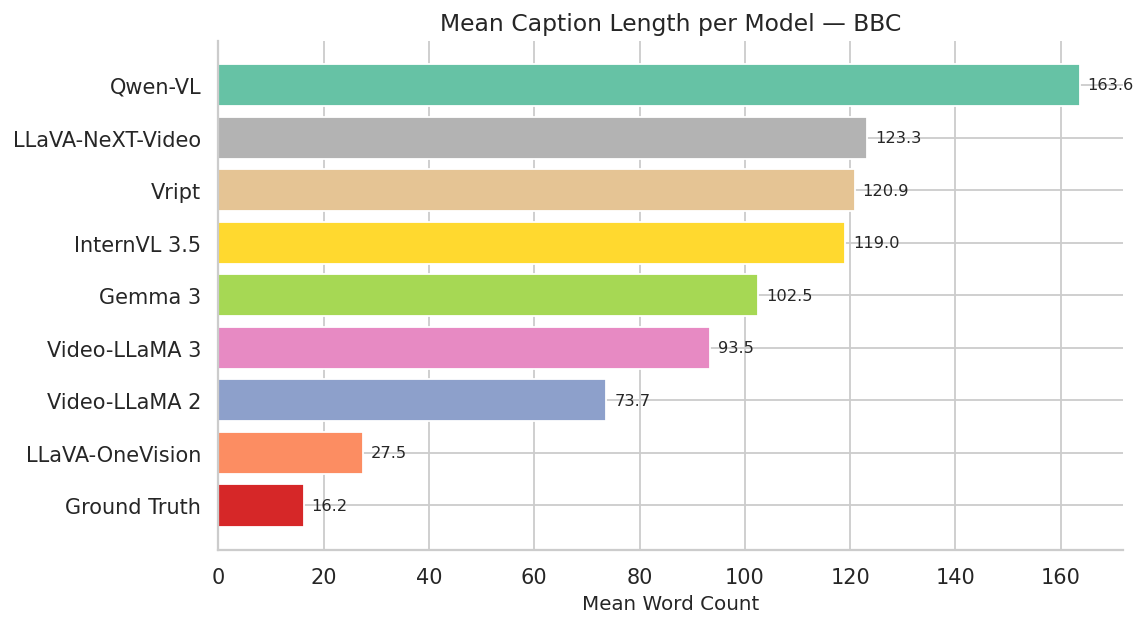}
  \caption{Mean caption length across models on the BBC dataset.}
  \label{fig:bbc_caption_len}
\end{figure}

Finally, we analyzed the length of the generated descriptions. Human reference descriptions are typically concise in both datasets, while all models tend to produce more verbose output. As shown in Figure~\ref{fig:bbc_caption_len}, models such as Vript and LLaVA-OneVision generate particularly long descriptions on the BBC dataset, while Gemma~3 and Qwen-VL produce moderately long outputs. This verbosity does not necessarily correlate with quality: models achieving the highest semantic similarity (Gemma~3) are not the most verbose, while some of the most verbose models (e.g., Vript) score lower on Text Similarity.

\begin{figure}[h!]
  \centering
  \includegraphics[width=0.8\linewidth]{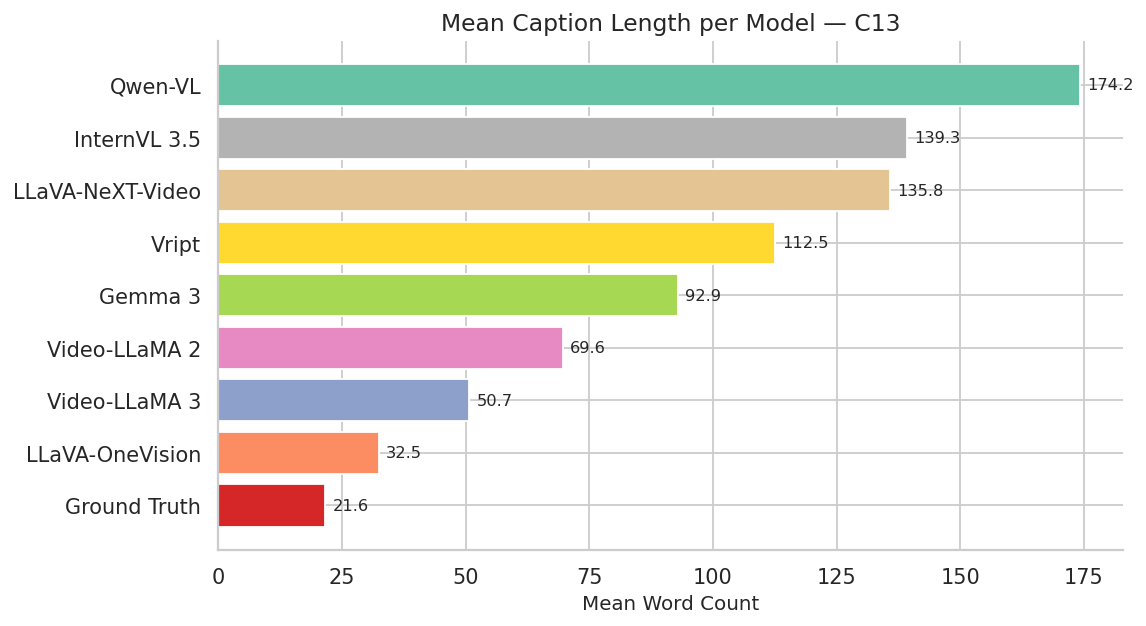}
  \caption{Mean caption length across models on the ChTV dataset.}
  \label{fig:chtv_caption_len}
\end{figure}

\subsection{Thematic Fidelity Score (TFS) and Entity Fidelity Score (EFS)}
\label{sec:tfs_results}
An important contribution of this work is the introduction of two new metrics for video description: the Thematic Fidelity Score (TFS) and the Entity Fidelity Score (EFS). The first one evaluates how faithfully a model's output caption preserves the thematic content of the ground-truth description, and the latter measures how effectively each model's output retrieves key named entities (persons, locations, organizations) present in the ground-truth description (see Section~\ref{sec:methodology}). 


Thus, Table~\ref{tab:tfs_efs_combined} presents the TFS and EFS scores for all models on both datasets. Regarding thematic fidelity, Gemma~3 achieves the highest TFS on the BBC dataset (0.912), while Qwen-VL slightly edges Gemma~3 on ChTV (0.865 vs.\ 0.861). All models achieve high TFS scores ($> 0.83$ on ChTV, $> 0.86$ on BBC), confirming that zero-shot thematic classification reliably captures the broad topic of the content. However, the TFS range is narrow across models---ChTV spans 0.839--0.865 (spread: 0.026) and BBC spans 0.862--0.912 (spread: 0.050) indicating that TFS alone has limited discriminative power for differentiating model quality. The BBC dataset shows slightly wider differentiation, likely due to the richer English language editorial descriptions that enable more nuanced thematic distinctions.

\begin{table}[h!]
  \centering
  \caption{Results of terms of  TFS and EFS values both datasets. Models sorted by BBC TFS. Bold values indicate the best score per column.}
  \label{tab:tfs_efs_combined}
  \begin{tabular}{lcc|cc}
    \hline
     & \multicolumn{2}{c|}{\textbf{ChTV}} & \multicolumn{2}{c}{\textbf{BBC}} \\
    Model & TFS  & EFS & TFS  & EFS \\
    \hline
    Gemma 3          & 0.861 & \textbf{0.059} & \textbf{0.912}  & \textbf{0.314} \\
    Qwen-VL          & \textbf{0.865} & 0.051 & 0.904 & 0.198 \\
    Video-LLaMA 2    & 0.852 & 0.007 & 0.895 & 0.053 \\
    LLaVA-OneVision  & 0.860 & 0.007 & 0.894 & 0.059 \\
    Video-LLaMA 3    & 0.849 & 0.002 & 0.893 & 0.138 \\
    InternVL 3.5     & 0.845 & 0.022 & 0.874 & 0.077 \\
    LLaVA-NeXT-Video & 0.849 & 0.010 & 0.871 & 0.022 \\
    Vript            & 0.839 & 0.008 & 0.862 & 0.047 \\
    \hline
  \end{tabular}
\end{table}


The EFS columns in Table~\ref{tab:tfs_efs_combined} reveal a striking contrast between the two datasets. On ChTV, EFS scores are extremely low across all models (0.002--0.059), reflecting the translated keyword style ground-truth that contains very few explicit named entities. Clips with zero ground-truth entities are excluded (assigned NaN), leaving few valuable samples and resulting in near-floor scores for all models. Gemma~3 leads marginally (0.059), followed by Qwen-VL (0.051), but the absolute values are too low to draw meaningful distinctions.

On the BBC dataset, in contrast, EFS scores show clear differentiation. Gemma~3 dominates with an EFS of 0.314, well ahead of Qwen-VL (0.198) and Video-LLaMA~3 (0.138). The English language BBC ground-truth descriptions contain richer named entity content, enabling substantially higher entity matching. Gemma~3 achieves the highest precision (0.280) and recall (0.446), with an average of 0.64 matched entities per clip. The remaining models cluster below 0.14, with LLaVA-NeXT-Video achieving the lowest EFS (0.022) on BBC.

The EFS results complement the TFS findings by highlighting a different dimension of caption quality. While TFS measures broad thematic alignment (where all models perform comparably), EFS evaluates the model's ability to identify and reproduce specific factual entities. On the BBC dataset, Gemma~3's dominance is even more pronounced than in TFS, suggesting that its advantage stems not only from thematic understanding but also from a strong capacity for entity recognition and grounding. The cross-dataset contrast underscores that entity-based evaluation requires entity-rich references to be meaningful, the ChTV EFS scores are essentially uninformative due to the sparse entity content of the translated ground-truth.

\subsection{Qualitative Analysis}
\label{sec:qualitative}

\subsubsection{Thematic Fidelity: Case Study (BBC)}
\label{sec:tfs_casestudy}

To illustrate how the Thematic Fidelity Score captures qualitative differences, we examine the BBC clip with the largest T-F1 gap between Gemma~3 (best overall on BBC, TFS~=~0.912) and LLaVA-NeXT-Video (worst overall on BBC, TFS~=~0.871). 

The ground-truth description \emph{``Migrants attempt to cross the Channel one day after 12 people died when their dinghy sank.''} spans three themes: \emph{International Affairs and Conflicts}, \emph{Crime and Justice}, and \emph{Accidents and Emergencies}.

\noindent Gemma~3 (TFS = 1.0, predicted themes: \emph{International Affairs and Conflicts, Crime and Justice, Accidents and Emergencies}):

\begin{quote}\small\itshape
``The video depicts a scene in Wimerueux, France, where a large group of migrants, many wearing red life jackets, are aboard a small, inflatable boat attempting to cross the English Channel. A French coast guard vessel is nearby, and on the beach, French police and coastguard are seen running towards the water. On-screen text reveals that this attempt occurred just a day after twelve migrants died making the same journey.''
\end{quote}

\noindent LLaVA-NeXT-Video (TFS = 0.53, predicted themes: \emph{Environment and Climate, Science and Technology, Arts/Culture/Entertainment, Natural Disasters and Weather}):

\begin{quote}\small\itshape
``In the video, we see a group of people on a boat in the ocean, with some of them wearing life jackets. The boat appears to be a tourist or sightseeing vessel, as it is carrying passengers and is relatively small compared to the size of the ocean. The passengers are looking towards the shore, possibly observing the landscape or waiting for an event to occur.''
\end{quote}

Gemma~3 correctly identifies all three themes present in the ground truth, achieving a perfect TFS of 1.0. It reproduces contextually accurate details such as the location (Wimerueux, France), the inflatable boat, the coast guard intervention, and the prior deaths. LLaVA-NeXT-Video, by contrast, misclassified the clip across four incorrect themes (Environment, Science, Arts/Entertainment, Natural Disasters), interpreting the migrant crossing as a ``tourist or sightseeing vessel''. This example illustrates a critical failure mode, models that rely heavily on surface level visual cues without integrating semantic context will consistently misclassify news segments that cover complex stories.

\subsubsection{Entity Fidelity: Case Studies (Both Datasets)}
\label{sec:efs_casestudy}

We present one representative video per dataset illustrating the largest EFS gap between the best and worst model.

\paragraph{BBC --- Gemma~3 (EFS~=~1.0) vs.\ LLaVA-NeXT-Video (EFS~=~0.0).}
The clip covers the story of Malcolm Metcalf, who took to the skies as part of completing 90 challenges before turning 90. Ground-truth: \emph{``Malcolm Metcalf took to the skies as he tries to complete 90 challenges before he turns 90.''}

\noindent Gemma~3 (EFS = 1.0, precision = 1.0, recall = 1.0):

\begin{quote}\small\itshape
``The video showcases an 89-year-old man, Malcolm Metcalf, taking control of a small airplane for a flight. The footage begins with preparations on the tarmac, then transitions to views from inside the cockpit as the plane takes off and soars over a rural landscape.''
\end{quote}

\noindent LLaVA-NeXT-Video (EFS = 0.0, precision = 0.0, recall = 0.0):

\begin{quote}\small\itshape
``In the video, we see two pilots in the cockpit of a small aircraft, likely a glider or a light aircraft, preparing for takeoff. One of the pilots is speaking to the camera, and the other is focused on the instruments and controls.''
\end{quote}

Gemma~3 correctly identifies the named individual (Malcolm Metcalf), his age (89), and the challenge context, achieving perfect entity precision and recall. LLaVA-NeXT-Video describes the same scene in purely visual terms, \emph{``two pilots in the cockpit''}, without inferring the protagonist's identity from available context (on-screen text, voiceover), yielding zero entity recall.

\paragraph{ChTV --- Gemma~3 (EFS~=~1.0) vs.\ Vript (EFS~=~0.0).}
The clip covers a news segment featuring a video call with a military officer discussing crime. Ground-truth: \emph{``Video call Lieutenant Daniel Medina talks about criminals and the characteristics of crimes and their types.''} One named individual (Lieutenant Daniel Medina) is present.

\noindent Gemma~3 (EFS = 1.0, precision = 1.0, recall = 1.0):

\begin{quote}\small\itshape
``The segment features a split screen. One side shows two women, one with a face shield, discussing a security camera video. The other side displays footage of a vehicle crashing through a barrier and into a roadway. A military officer, Lieutenant Medina, is shown in a separate video clip, providing commentary.''
\end{quote}

\noindent Vript (EFS = 0.0, precision = 0.0, recall = 0.0):
\begin{quote}\small\itshape
``The video features a news segment with a split-screen layout. On the left side, there's a man in a military uniform, and on the right side, there's a woman wearing a pink top. They are both seated in front of a background that appears to be a news studio.''
\end{quote}

Gemma~3 correctly names the military officer (Lieutenant Medina), achieving perfect entity scores. Vript describes the same individual only as ``a man in a military uniform,'' without identifying the person by name, yielding zero entity recall. This cross-dataset contrast reinforces the aggregated EFS findings: on the BBC dataset, where descriptions are in English and models generate English captions, entity recall is substantially higher overall; on ChTV, the translated keyword style ground-truth makes entity matching inherently difficult, but the qualitative case shows that Gemma~3 can still extract named entities from on-screen text even in the Chilean news domain.

\subsection{Methodological Discussion}

This work makes three methodological contributions. First, we proposed two benchmark datasets for news video captioning that encompass different languages, editorial styles, and scales, providing a more robust evaluation setting than a single dataset study. The cross-dataset analysis proved particularly informative: the BBC dataset, with its richer English-language editorial descriptions, yielded wider performance spreads and sharper model differentiation (e.g., Text Similarity range of 0.209 on BBC vs.\ 0.132 on ChTV), while the ChTV dataset highlighted the challenges of evaluating against translated, tag-based annotations.

Second, the two proposed fidelity metrics TFS and EFS offer novel evaluation dimensions that go beyond surface-level similarity. Crucially, the experimental analysis demonstrated that standard metrics---including METEOR, ROUGE-L, CLIPScore, and BERTScore---are ill-suited for differentiating model quality in news video captioning, due to surface-form dependence, static-frame insensitivity, and function-word inflation, respectively. TFS and EFS address these gaps directly. TFS quantifies how well a caption preserves the thematic structure of the source content through zero-shot DeBERTa classification across 15 candidate themes, while EFS measures named-entity coverage via spaCy NER and fuzzy matching as the harmonic mean of entity precision and recall (EFS). Both metrics exposed model behaviours not captured by standard measures: for instance, all models achieved high TFS scores ($> 0.83$), with a relatively narrow spread (0.026 on ChTV, 0.050 on BBC), indicating that thematic classification alone has limited discriminative power; meanwhile, EFS on the ChTV dataset revealed near-floor scores (${\sim}$0.002--0.059) due to entity-sparse ground-truth, underscoring that metric informativeness is inherently dependent on the richness of the reference data.

Third, the shuffled-pairs validation demonstrated the discriminative power of the Text Similarity metric: Gemma~3 showed clear separation between original and shuffled distributions on both datasets, whereas lower-performing models (LLaVA-NeXT-Video on BBC, Video-LLaMA~3 on ChTV) exhibited substantially less separation. This test also highlighted an important nuance: low scores may reflect a mismatch between a model's descriptive style and the nature of the ground-truth annotations rather than a fundamental limitation in the model's capabilities.

%% file: 05_conclusions.tex
\label{sec:conclusions}

\section{Conclusions}

This paper presented a comprehensive comparative study of eight state-of-the-art Video Large Language Models (Gemma~3, Vript, InternVL~3.5, Video-LLaMA~2, Video-LLaMA~3, LLaVA-NeXT-Video, LLaVA-OneVision, and Qwen-VL) for the task of automatic video description in the challenging domain of broadcast television news. The evaluation was conducted on two complementary benchmark datasets a Chilean TV news corpus (ChTV, ${\sim}$1,345 clips) and a BBC News corpus (${\sim}$9,838 clips) using a multi-dimensional suite of metrics that spans n-gram overlap (METEOR, ROUGE-L), semantic similarity (BERTScore, CLIPScore, Text Similarity, MRR), and two novel fidelity measures introduced in this work: the Thematic Fidelity Score (TFS) and Entity Fidelity Score (EFS).

\subsection{Key Findings}

Our primary finding is the consistent superiority of Gemma~3 as the most balanced performer across both datasets and most evaluation dimensions. Gemma~3 achieved the highest Text Similarity (0.439 on BBC, 0.311 on ChTV), the highest METEOR scores (0.202 on BBC, 0.128 on ChTV), the highest CLIPScore on both datasets (0.364 on BBC, 0.339 on ChTV), and the top MRR on both datasets, confirming that it generates descriptions most semantically aligned with the editorial references and that it consistently ranks first among all models for any given video-clip.

This dominance extends to the two proposed fidelity metrics. On TFS, Gemma~3 attained the highest TFS on the BBC dataset (0.912), while on ChTV Qwen-VL marginally leads (0.865 vs.\ 0.861). On EFS, Gemma~3 led decisively on the BBC dataset (EFS~=~0.314), demonstrating a strong capacity for recognizing and reproducing specific named entities with an average of 0.64 matched entities per clip and a recall of 0.446, far exceeding all other models. Qwen-VL emerged as a consistent runner-up across Text Similarity, TFS, and EFS, confirming its position as the second-strongest model for news video captioning.

However, no single model dominated uniformly across all metrics. LLaVA-OneVision achieved the best ROUGE-L (0.120 on BBC, 0.082 on ChTV) and BERTScore F1 (0.836 on BBC, 0.791 on ChTV). This divergence reveals that BERTScore captures token-level semantic alignment, while metrics such as Text Similarity and TFS better reflect contextual and thematic alignment with the editorial references. The complementary nature of these metrics reinforces the need for a multi-dimensional evaluation framework in video captioning research.

\subsection{Limitations}

Several limitations should be noted. First, the ground-truth annotations in both datasets were created for editorial indexing, not as literal visual descriptions. This annotation bias favors models that align with the editorial style (Gemma~3, Qwen-VL) and may undervalue models with strong visual grounding but different descriptive strategies. Second, n-gram metrics (METEOR, ROUGE-L) remain limited for assessing abstract descriptions, as evidenced by the uniformly low scores across all models. Third, the TFS metric showed limited discriminative power due to the narrow range of scores across models (spread of 0.026 on ChTV and 0.050 on BBC), suggesting that zero-shot thematic classification alone may not sufficiently differentiate model quality. Fourth, the E-F1 metric proved uninformative on the ChTV dataset due to the low entity density in the translated ground-truth (most clips yielding NaN), suggesting that entity-based evaluation requires entity-rich references to be meaningful. Finally, inferences were performed with default configurations and prompts for all models; task-specific prompting or fine-tuning could alter the relative rankings.

\subsection{Future Work}

Based on these findings, we identify several directions for future research.

First, a natural next step is to refine the ground-truth annotations. Creating richer, purpose-built descriptions that combine literal visual narration with editorial context would enable more informative evaluation across all metrics and reduce the annotation bias observed in this study. Alternatively, having models produce structured outputs such as thematic categories paired with timestamps could better align with the functional needs of media archives.

Second, the complementary strengths observed across models suggest that ensemble or hybrid approaches merit investigation. For instance, combining Gemma~3's thematic and entity grounding with LLaVA-OneVision's visual alignment capabilities could yield descriptions that excel across both semantic and perceptual dimensions.

Third, while TFS and EFS represent a step towards domain specific evaluation, further work is needed to develop metrics that assess journalistic qualities such as narrative coherence, factual accuracy, and informational completeness. Extending TFS and EFS to other domains beyond news such as sports or documentary content would also test the generality of these metrics.

Finally, deploying these models in real-world production environments, where journalists, editors, and archivists can interact with and provide feedback on the generated descriptions, would offer valuable insights into practical utility and guide iterative improvements towards more effective automated content management systems.